\documentclass[sigconf]{acmart}

\AtBeginDocument{%
  }

\setcopyright{acmlicensed}
\copyrightyear{2024}
\acmYear{2024}
\setcopyright{rightsretained}
\acmConference[GECCO '24 Companion]{Genetic and Evolutionary Computation Conference}{July 14--18, 2024}{Melbourne, VIC, Australia}
\acmBooktitle{Genetic and Evolutionary Computation Conference (GECCO '24 Companion), July 14--18, 2024, Melbourne, VIC, Australia}
\acmDOI{10.1145/3638530.3654394}
\acmISBN{979-8-4007-0495-6/24/07}

\usepackage{graphicx}
\usepackage{subcaption}
\usepackage{pdfpages}
\usepackage{algorithm,algpseudocode}
\usepackage{booktabs}

\begin{document}

%%
%% The "title" command has an optional parameter,
%% allowing the author to define a "short title" to be used in page headers.
\title{Tensorized Ant Colony Optimization
for GPU Acceleration}

%%
%% The "author" command and its associated commands are used to define
%% the authors and their affiliations.
%% Of note is the shared affiliation of the first two authors, and the
%% "authornote" and "authornotemark" commands
%% used to denote shared contribution to the research.

\author{Luming Yang}
\email{skylynfluke@gmail.com}
\affiliation{%
% \department{Department of Computer Science and Engineering}
  \institution{Southern University of Science and Technology}
  \city{Shenzhen}
  \state{Guangdong}
  \country{China}
}

\author{Tao Jiang}
\email{jiangt_97@163.com}
\affiliation{%
% \department{Department of Computer Science and Engineering}
  \institution{Southern University of Science and Technology}
  \city{Shenzhen}
  \state{Guangdong}
  \country{China}
}

\author{Ran Cheng}
\authornote{Corresponding Author}
\email{ranchengcn@gmail.com}
\affiliation{%
% \department{Department of Computer Science and Engineering}
  \institution{Southern University of Science and Technology}
  \city{Shenzhen}
  \state{Guangdong}
  \country{China}
}

%%
%% By default, the full list of authors will be used in the page
%% headers. Often, this list is too long, and will overlap
%% other information printed in the page headers. This command allows
%% the author to define a more concise list
%% of authors' names for this purpose.
\renewcommand{\shortauthors}{Yang, Jiang and Cheng}

%%
%% The abstract is a short summary of the work to be presented in the
%% article.
\begin{abstract}

Ant Colony Optimization (ACO) is renowned for its effectiveness in solving Traveling Salesman Problems, yet it faces computational challenges in CPU-based environments, particularly with large-scale instances.
In response, we introduce a Tensorized Ant Colony Optimization (TensorACO) to utilize the advancements of GPU acceleration.
As the core, TensorACO fully transforms ant system and ant path into tensor forms, a process we refer to as \emph{tensorization}.
For the tensorization of ant system, we propose a preprocessing method to reduce the computational overhead by calculating the probability transition matrix.
In the tensorization of ant path, we propose an index mapping method to accelerate the update of pheromone matrix by replacing the mechanism of sequential path update with parallel matrix operations.
Additionally, we introduce an Adaptive Independent Roulette (AdaIR) method to overcome the challenges of parallelizing ACO's selection mechanism on GPUs.
Comprehensive experiments demonstrate the superior performance of TensorACO achieving up to 1921$\times$ speedup over standard ACO.
Moreover, the AdaIR method further improves TensorACO's convergence speed by 80\% and solution quality by 2\%. Source codes are available at \href{https://github.com/EMI-Group/tensoraco}{https://github.com/EMI-Group/tensoraco}.
\end{abstract}
%%
%% The code below is generated by the tool at http://dl.acm.org/ccs.cfm.
%% Please copy and paste the code instead of the example below.
%%
\begin{CCSXML}
<ccs2012>
   <concept>
       <concept_id>10010147.10010169.10010170.10010173</concept_id>
       <concept_desc>Computing methodologies~Vector / streaming algorithms</concept_desc>
       <concept_significance>500</concept_significance>
       </concept>
   <concept>
       <concept_id>10010147.10010178.10010205.10010207</concept_id>
       <concept_desc>Computing methodologies~Discrete space search</concept_desc>
       <concept_significance>500</concept_significance>
       </concept>
   <concept>
       <concept_id>10002950.10003624.10003625.10003630</concept_id>
       <concept_desc>Mathematics of computing~Combinatorial optimization</concept_desc>
       <concept_significance>500</concept_significance>
       </concept>
 </ccs2012>
\end{CCSXML}

\ccsdesc[500]{Computing methodologies~Vector / streaming algorithms}
\ccsdesc[500]{Computing methodologies~Discrete space search}
\ccsdesc[500]{Mathematics of computing~Combinatorial optimization}

\keywords{Tensorization, ACO, GPU, Adaptive Independent Roulette, TSP}

% \received{20 February 2007}
% \received[revised]{12 March 2009}
% \received[accepted]{5 June 2009}

%%
%% This command processes the author and affiliation and title
%% information and builds the first part of the formatted document.
\maketitle

\section{Introduction}

The Traveling Salesman Problem (TSP), a central challenge in combinatorial optimization, is characterized by the theoretical complexity and practical relevance in routing and logistics \cite{Reinelt2003, Juenger1995}. As a classic problem, large-scale TSP instances require computational resources that increase exponentially with the size of the problem. Ant Colony Optimization (ACO) \cite{Dorigo1996}, a classic metaheuristic algorithm inspired by the foraging behavior of ants, has emerged as a particularly effective tool for TSP. However, the computational bottleneck becomes apparent with the rapid advancements in problem scale. 

The performance of GPU within a tensorized computational framework is recognized in processing massively parallel computations \cite{Feng2023}.  
The advancement of GPU architectures and the emergence of tensorized computing provide a unique opportunity to transform traditional CPU-based ACO models. The significant challenges of GPU-accelerated ACO are the intrinsic complexity and sequential nature of traditional computational logic \cite{Dzalbs2020}.  A paradigm shift towards a GPU-optimized framework is envisaged to enhance the performance of ACO significantly, especially in addressing complex large-scale optimization challenges.

The JAX framework \cite{jax2018github} enhances high-performance numerical computing with hardware acceleration significantly.
Building upon JAX, pioneering platforms like EvoJAX \cite{evojax}, evosax \cite{evosax}, and EvoX \cite{huang2023evox} have emerged, providing efficient tools for evolutionary computation for advanced hardware acceleration. However, a complete implementation of scalable ACO remains absent.
To bridge this gap, we propose the Tensorized Ant Colony Optimization (TensorACO) for GPU Acceleration.

The main contributions of this paper are summarized below.

\begin{itemize}
\item We propose an ant system tensorization method which integrates heuristic values and the pheromone matrices to compute the probability matrix, enabling preprocessing while achieving acceleration using function mapping.

\item We propose an ant path tensorization method which parallelizes the serial processing of ant path solutions in the pheromone matrix computation and reduces redundant path cost calculations through an index mapping method.

\item We propose an Adaptive Independent Roulette (AdaIR) selection mechanism which improves performance while retaining the benefits of parallelization by dynamically adjusting the tendency of ants to select cities with higher probabilities. 
\end{itemize}

\begin{figure*}[htbp]
    	\centering
    	\includegraphics[width=0.8\linewidth]{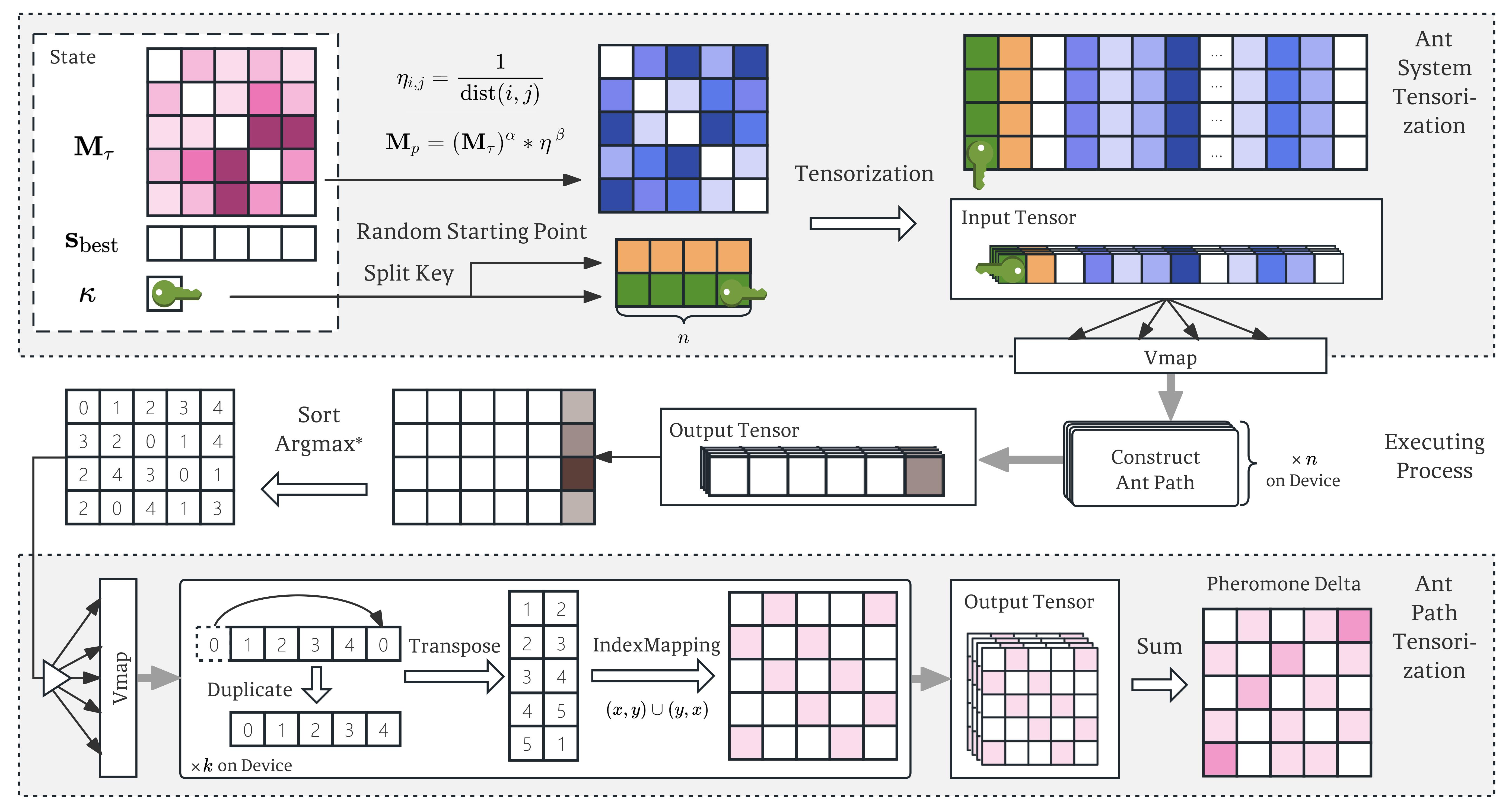}
        \caption{ Schematic overview of TensorACO. The workflow comprises two main components: \emph{ant system tensorization} and \emph{ant path tensorization}. The matrix of squares represents the pheromone matrix or probability transition matrix, and the color depth represents the value. The key represents $\kappa$. Multiple arrows represent the function mapping, and multiple overlay sets pointed by the arrows can be parallelized on device. }
        \label{Figure model}
\end{figure*}

\section{Method}

 The schematic overview of TensorACO is shown in Figure \ref{Figure model}. 
 TensorACO encompasses several key components: Preprocessing, Ant System Tensorization, Ant Path Tensorization, and an Adaptive Independent Roulette selection mechanism. 
 The symbol $n$ represents the number of cities in TSP, and $m$ represents the number of ants in ACO.

\subsection{Preprocessing}

We designed a preprocessing method that constructs pheromone probability transition matrix $\mathbf{M}_p$ beforehand, replacing the computation of pheromone concentration $\mathbf{M}_\tau$ in each iteration. The calculation of $\mathbf{M}_p$ can be described as follows:
\begin{equation}
\mathbf{M}_p = \text{RowNormalized}(\mathbf{M}_{i,j}) = \frac{(\mathbf{M}_\tau^\alpha \times \eta^\beta)_{i,j}}{\sum_{j=1}^{n} (\mathbf{M}_\tau^\alpha \times \eta^\beta)_{i,j}},
\end{equation}
where $\text{RowNormalized}(\mathbf{M}_{i,j})$ is the normalization method, $\eta_{i,j}$ is the heuristic function value from path $i$ to $j$, 
$\alpha$ and $\beta$ are parameters controlling the importance of pheromone and heuristic information. 

\subsection{Tensorization of Ant System}

We proposed a tensorized method which utilizes functional programming and treats the ant movement as a function. The tensorization of ant system can be described as follows:

$\mathbf{Initialization}$: The initial solutions are generated randomly as a tensor, which is formulated as 
$
\mathbf{T}_{\text{start}} = \begin{bmatrix}
{v}_{1},\ {v}_{2},\ \cdots,\ {v}_{n}
\end{bmatrix}.
$
$\mathbf{T}_{\text{start}}$ contains randomly generated integer $v_i \in \mathbb{Z}^+ \cap [0, m), \ \forall i = 1, 2, \ldots, n.$

$\mathbf{Tensorization}$:
For the transition probability matrix $\mathbf{M}_p$, we reshape it into a one-dimensional tensor $\mathbf{v}_p \in \mathbb{R}^{m^2}$ and then perform a duplication operation, resulting in a tensor $\mathbf{T}_p \in \mathbb{R}^{n\times m^2}$. By merging these three tensors, we obtain a tensor
$\mathbf{T}_{\text{input}} \in \mathbb{R}^{n \times (2+m^2)}.$
Lastly, the tensor is reshaped along the ant dimension to complete the tensorization of the ant colony.

\subsection{Tensorization of Ant Path}

We specifically developed a method for the tensorization of ant path, catering to ACO variants that require referencing multiple paths for updates. Upon obtaining the solution set $\mathbf{T}_{\text{output}}$, we sort it and select the top $k$ paths. The selected ant path solution set is described as 
$\mathcal{P}= \begin{bmatrix}
{\mathbf{s}_1},\ {\mathbf{s}_2},\ ,\cdots,\ {\mathbf{s}_l},\ \cdots,\ {\mathbf{s}_k}
\end{bmatrix}$
, which is the input in the function mapping process, mapped along dimension $\mathbf{s}_l$.

During the processing of the solution, we duplicate each path twice with a one-unit shift to construct the index matrix $\mathbf{I}$:
\begin{equation}
\mathbf{I} = \left(\begin{bmatrix}
 p_1 & p_2 & \cdots  & p_{m-1} & p_m\\
 p_m & p_1 & p_2 & \cdots & p_{m-1}
\end{bmatrix}\right)^T.
\end{equation}
An increment matrix is then generated as follows:
\begin{equation}
\mathbf{A}_{i,j}=\left\{\begin{matrix}\frac{1}{f(\mathbf{s}_l)}, & \exists k,\ \mathbf{e}_k\mathbf{I}=[i,j] \lor \mathbf{e}_k\mathbf{I}=[j,i] \\0, & \text{Otherwise}\end{matrix} \right.,
\end{equation}
where $f$ is the fitness function, $\{\mathbf{e}_1,\cdots,\mathbf{e}_m\}$ are the standard basis vectors in $\mathbb{R}^m$.
By combining these matrices, we obtain the final pheromone increment 
${\Delta\mathbf{M}_{\tau_{i,j}}}=\Sigma_{\mathcal{P}}\mathbf{A}_{i,j}$.

\subsection{Adaptive Selection Mechanism}

We designed an Adaptive Independent Roulette (AdaIR) method based on the Independent Roulette (IR) method, enabling the adaptive selection of new solutions.
The IR method replaces the Roulette Wheel (RW) process with a parallel-compatible alternative. This method introduces a random deviate, a randomly generated tensor $\textbf{r} \in \mathbb{R}^m$. The process of selecting next city is abstracted as $v = \arg \max (\textbf{r} \odot \textbf{p}_u)$, where $\odot$ is element-wise product.

The AdaIR leverages a \textit{learning rate} hyperparameter (denoted as $\gamma$) to accelerate convergence.
\begin{equation}
\mathbf{r}'=(r_1^{\gamma}, r_2^{\gamma}, r_3^{\gamma}, \ldots, r_m^{\gamma}),
\end{equation}
where $\mathbf{r}$ signifies the randomly generated tensor, with its elements $r_i$ uniformly distributed within the interval $(0,1)$, i.e., $X \sim \mathbf{U}(0,1)$.
Let $Y$ represent the transformed variable after exponentiation by $\gamma$, such that:
$ Y = X^{\gamma}. $ Next, $F_Y(y) = P(X \leq y^{\frac{1}{\gamma}}).$
The Probability Density Function (PDF) of $Y$ can be obtained:
\begin{equation}
f_Y(y)=\begin{cases}
\frac{1}{\gamma}y^{\frac{1-\gamma}{\gamma}}, & 0 < y < 1 \\
0, & \text{Otherwise}
\end{cases}.
\end{equation}
The distribution's shape varies with $\gamma$: for $\gamma > 1$, the distribution skews towards smaller values, implying a greater degree of randomization in the adapted probability tensor. 

In AdaIR, we adopt the Cosine Annealing learning rate schedule as the adjustment strategy. The $\gamma$ hyperparameter is dynamically tuned during the iteration process, making the convergence process more rational and efficient.

\section{Experiments}

 All experiments were conducted on a uniform platform and compatible with the EvoX \cite{huang2023evox} framework. (CPU: Intel Core i7-10750H @ 2.60GHz; GPU: NVIDIA A100 Tensor Core; Python 3.9; CUDA 12.2; JAX 0.4.23). The dataset employed in this experiment is the TSPLIB dataset \cite{reinelt1991tsplib}.

\subsection{Acceleration Performance}

To evaluate the acceleration effect of TensorACO, tests were conducted across varying population sizes (from $10^0$ to $10^4$) and city scales (using six representative TSPLIB instances: \textit{u159, pcb442, p654, u724, pcb1173,} and \textit{pr2392}). The algorithm was executed 1000 times with $m=n$ and $k=n/10$, and the average time per iteration was calculated over 5 runs with independent keys.

\begin{figure}[htbp]
  \centering
  \includegraphics[width=1\linewidth]{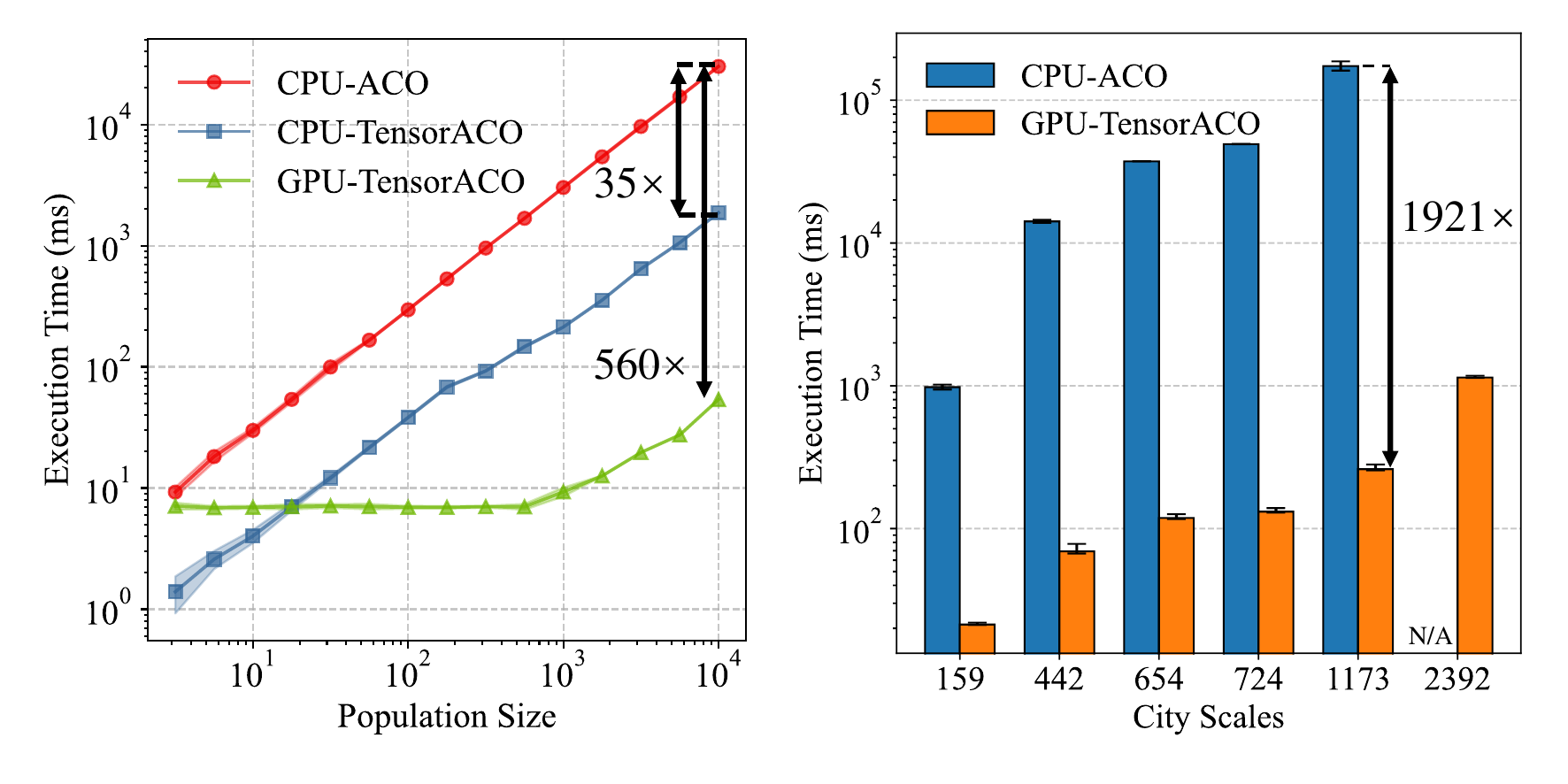}
  \captionsetup{skip=2pt}
  \setlength{\belowcaptionskip}{-10pt}
  \vspace{-5pt}
  \caption{Runtime over scaling population size and scaling city size. \emph{Note}: No data is available at the city scale of 2392 as the runtime of CPU-ACO exceeded the tolerance range.}
  \vspace{-10pt}
  \label{cpu-cpu-gpu}
\end{figure}

The results in Figure \ref{cpu-cpu-gpu} demonstrate significant speedups for TensorACO on both CPU and GPU platforms compared to CPU-ACO. CPU-TensorACO achieved up to 35$\times$ speedup, while GPU-TensorACO reached up to 560$\times$ acceleration. Across city scales, GPU-TensorACO consistently showed lower execution times, achieving a remarkable 1921$\times$ speedup for \textit{pcb1173}. These results underscore the substantial benefits of tensorization and GPU acceleration, particularly for large-scale optimization problems with higher computational demands.

\begin{figure*}[htbp]
\centering
\subfloat[CPU-TensorACO]{\includegraphics[width=0.33\linewidth]{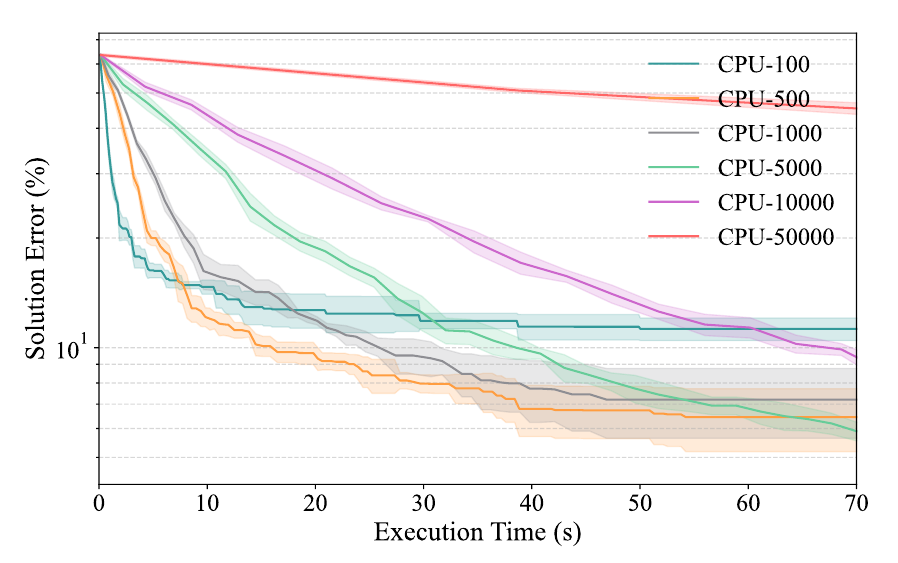}\label{subfig:cpu-convergence}}
\hfill
\subfloat[GPU-TensorACO]{\includegraphics[width=0.33\linewidth]{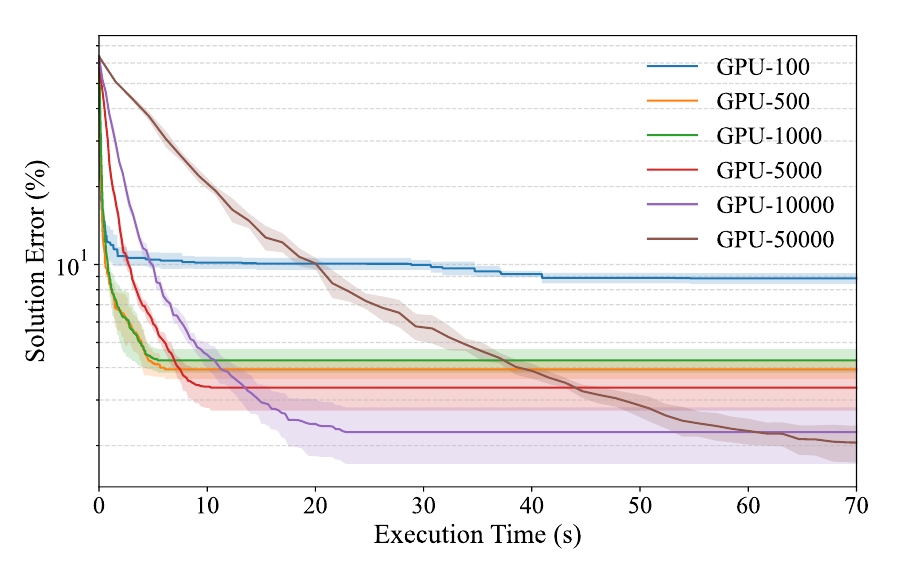}\label{subfig:gpu-convergence}}
\hfill
\subfloat[Comparison Result]{\includegraphics[width=0.33\linewidth]{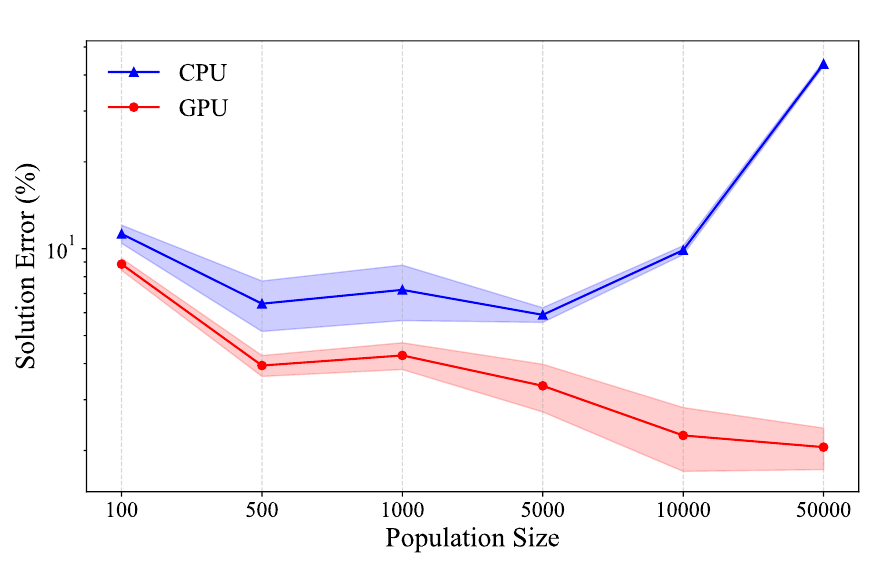}\label{subfig:cpu-gpu-performance}}
\captionsetup{skip=2pt}
\setlength{\belowcaptionskip}{-10pt}
\caption{ Solution error for CPU-TensorACO and GPU-TensorACO with varying population sizes. (a) and (b): Convergence curve over runtime; (c): Final quality obtained by two models for 70s. On GPUs, a larger population size helps to achieve better results.}
\vspace{-2pt}
\label{fig:combined}
\end{figure*}

\subsection{Convergence Performance}

An experiment was conducted to assess the convergence performance of TensorACO on CPU and GPU platforms, using instances with city sizes ranging from 100 to 50,000. With a fixed time limit of 70 seconds and five independent runs per instance, the solution errors (percentage difference from a specific solution) were evaluated. The results, as shown in Figure \ref{fig:combined}, revealed the influence of colony size on TensorACO's convergence speed and accuracy. Larger populations generally slowed convergence but improved solution accuracy on both CPU and GPU, indicating broader searches yielding more precise outcomes. However, at population sizes of 10,000 and 50,000, CPU performance was suboptimal, likely due to convergence challenges in expansive search spaces.

In contrast, GPU-accelerated TensorACO leveraged the GPU's computational prowess, confirming that larger populations can enhance performance given sufficient computational speed. Furthermore, allocating more computational resources and increasing ant colony size are hypothesized to significantly enhance the solution quality after convergence, as the GPU's superior computational efficiency enables more comprehensive searches within the given time constraints.

\subsection{Ablation Study on AdaIR}

\subsubsection{Visualization} 

Visualization analyses on instances with 100 and 422 cities revealed that AdaIR initially shifts the maximum probability laterally to higher values, gradually converging to the same probabilities as the RW method. 

In typical TSP scenarios, we sampled probability tensors in the path construction process. We only consider the city with the highest original probability $p_\text{max}$. The AdaIR method is then applied to these tensors, with $\gamma$ hyperparameter consistently set to 1.5.  We performed \(m \times 10,000\) Bernoulli trials on the tensors, yielding \(\hat{p}_{\text{max}}'\). 
In the initial few generations, the probability was shifted laterally to a higher level and gradually became the same as RW.

\begin{figure}[htbp]
  \centering
  \begin{subfigure}[b]{0.35\linewidth}
    \centering
    \includegraphics[width=\linewidth]{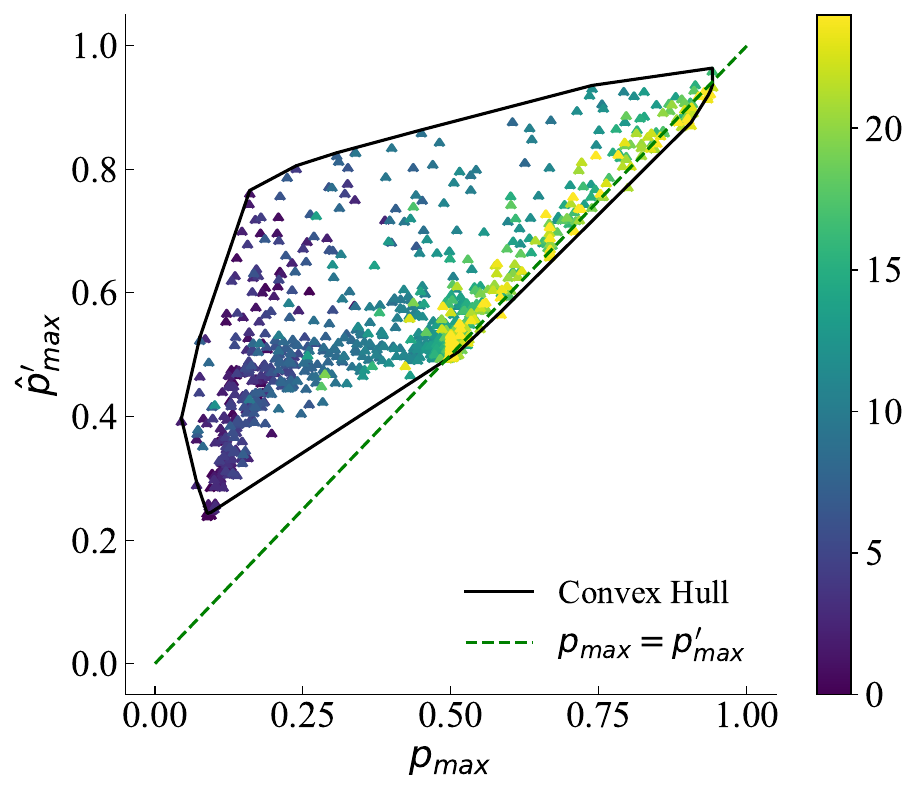}
    \caption{$m=442,\ \gamma=1$}
  \end{subfigure}
  \hfill
  \begin{subfigure}[b]{0.31\linewidth}
    \centering
    \includegraphics[width=\linewidth]{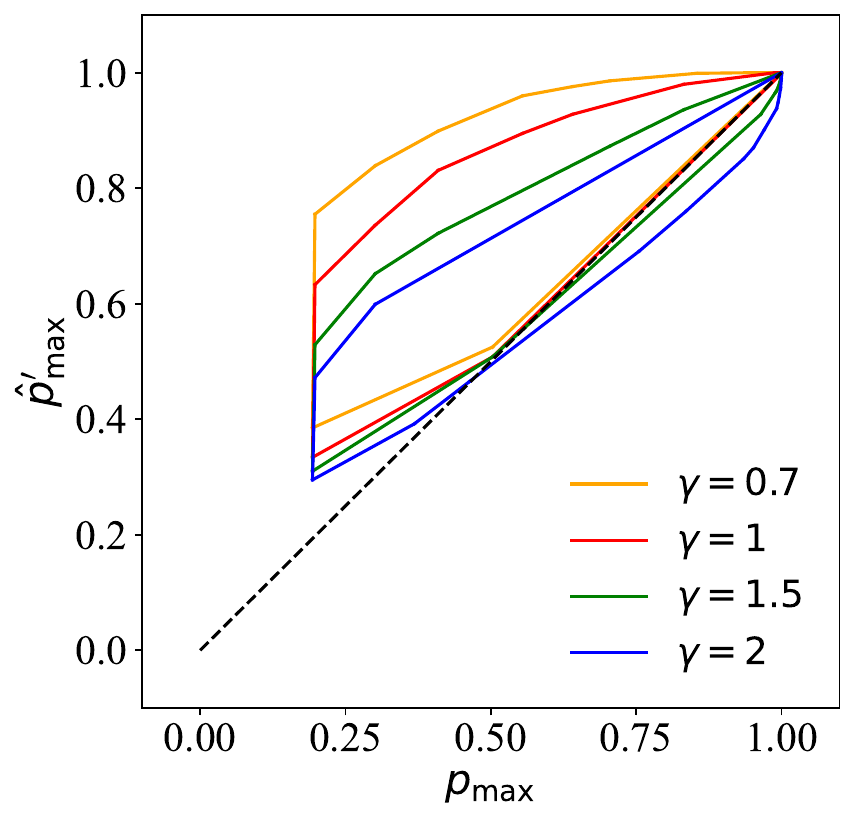}
    \caption{$m=100$}
  \end{subfigure}
  \hfill
  \begin{subfigure}[b]{0.31\linewidth}
    \centering
    \includegraphics[width=\linewidth]{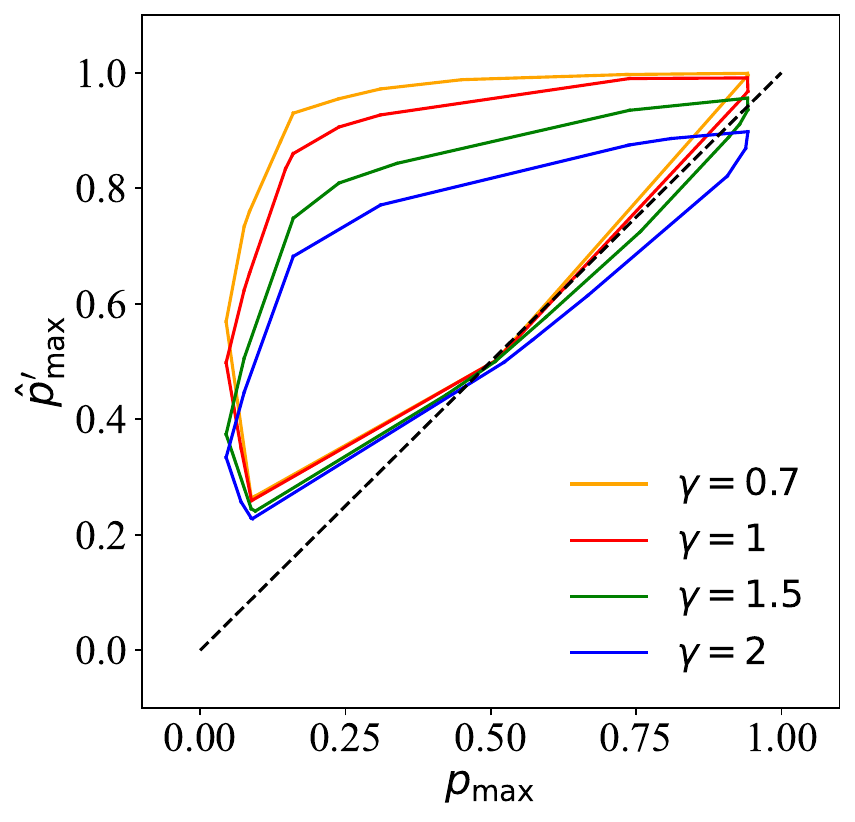}
    \caption{$m=442$}
  \end{subfigure}
  \captionsetup{skip=5pt}
  \caption{ Relationship between RW probability $p_\text{max}$ and shifted probability $\hat{p}_{\text{max}}'$ using AdaIR. The gradient color represents the variation across iterations. $\hat{p}_{\text{max}}$ and $\hat{p}_{\text{max}}'$ tend to align within the first few iterations. (c): As $\gamma$ varies, the convex hull alters, guiding the directional shift in ants' selection tendencies.}
  \vspace{-5pt}
  \label{IRP}
\end{figure}

Figure \ref{IRP} shows the convex hull probability ranges for different learning rates ($\gamma$) demonstrating AdaIR's adaptability, with higher $\gamma$ values encouraging more diverse exploration. The implementation of the learning rate $\gamma$ facilitates a more dynamic range of the IR convex hull's expansion and contraction, making the IR method more adaptable to the search process's dynamic needs and improving the balance between exploration and exploitation.

\subsubsection{Acceleration}
\begin{figure}[htbp]
  \centering
  \includegraphics[width=\linewidth]{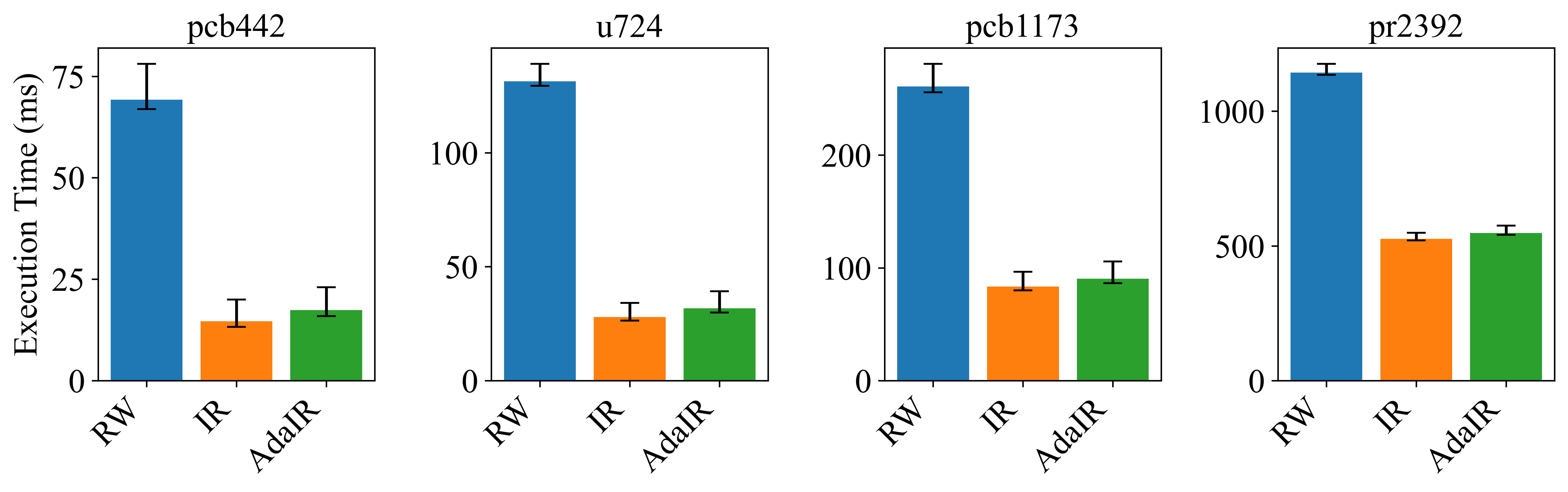}
  \captionsetup{skip=5pt}
  \vspace{-5pt}
  \caption{ Runtime for RW, IR, and AdaIR over varying city scales. AdaIR achieves speedup ranging from 2.12$\times$ to 4.62$\times$ against RW, with only a slight time difference from IR.}
  \vspace{-5pt}
  \label{time-ir}
\end{figure}

We compared RW, IR, and AdaIR methods using the same settings, testing the algorithm runtime per iteration on six instances with different city numbers.
Figure \ref{time-ir}, shows AdaIR demonstrates a speedup of more than 2.12$\times$ over RW. This highlights AdaIR’s feasibility as a competitive alternative to IR in terms of execution speed while offering the additional advantage of adaptive solution quality enhancements.

\subsubsection{Convergence}
\begin{figure}[htbp]
  \centering
  \includegraphics[width=\linewidth]{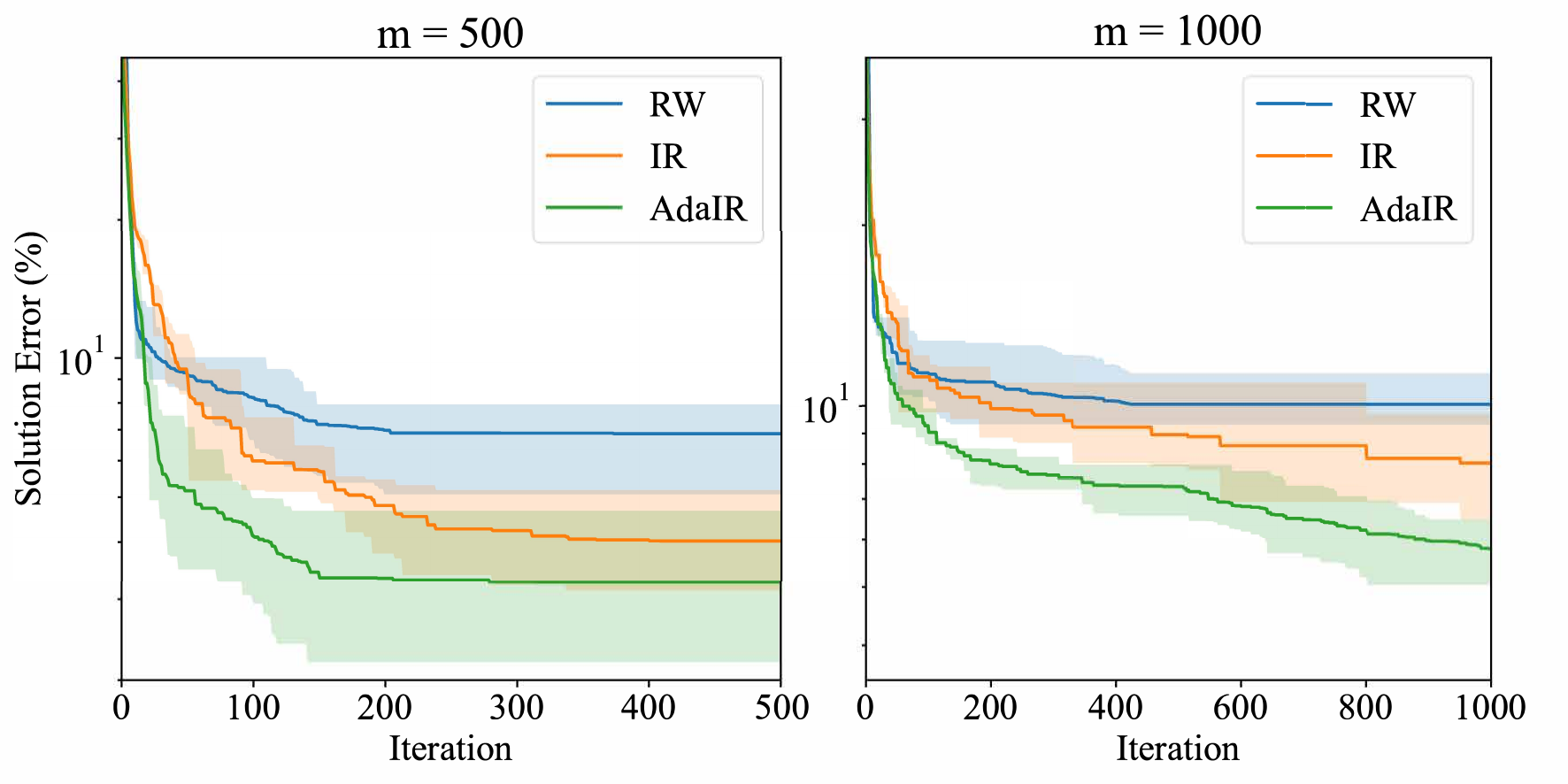}
  \captionsetup{skip=2pt}
  \vspace{-5pt}
  \caption{ Convergence curves of RW, IR, and AdaIR over iterations. AdaIR outperforms RW and IR both in terms of convergence speed and solution error. }
  \vspace{-5pt}
  \label{convergence}
\end{figure}

We experimented with RW, IR, and AdaIR methods over 1000 iterations, repeated five times.
Figure \ref{convergence} documents both the range and mean quality across trials. For $m=500$, AdaIR converges in 165 generations, outperforming RW, which takes 310 generations, and IR in terms of solution quality. 
This enhanced speed is due to AdaIR’s power-law \textit{learning rate}, this adaptability allows comprehensive exploration and higher convergence accuracy, reducing stagnation risk and potentially improving the likelihood of finding globally optimal solutions. The result underscores AdaIR's effectiveness in complex optimization challenges.

\section{Conclusion}

This study has introduced Tensorized Ant Colony Optimization (TensorACO), which leverages the tensorization method alongside GPU’s parallel processing capabilities to overcome the computational challenges faced by traditional ACO algorithms. 
At the core of TensorACO is the tensorization of ant system and ant path.
Furthermore, the Adaptive Independent Roulette (AdaIR) method in TensorACO enhances the algorithm's convergence speed while maintaining the quality of solutions. 
The efficacy of TensorACO is evident through significant computational speedups and improved scalability, proving its potential as an effective solution for large-scale Traveling Salesman Problem (TSP) instances.

Future work could focus on extending the tensorized framework and exploring AdaIR.
The potential of TensorACO in various other challenging optimization problems is also worth investigating.

%This study presents the development and evaluation of a Tensorized Ant Colony Optimization (TensorACO) for GPU acceleration. The tensorization method, combined with the GPU’s parallel processing capabilities, addresses the inherent computational bottlenecks of traditional ACO. Combined with the GPU’s parallel processing capabilities, TensorACO achieves an acceleration of approximately 2000 times compared to the standard ACO. Moreover, the Adaptive Independent Roulette (AdaIR) method introduced in this study further enhances the convergence speed of the algorithm, without sacrificing the solution qualities. The adaptive nature of this mechanism enables the algorithm to dynamically adjust its search strategy, striking a balance between exploration and exploitation. TensorACO demonstrates its efficacy through substantial computational speedups and scalability, making it a viable solution to large-scale TSP instances.

%%
%% The next two lines define the bibliography style to be used, and
%% the bibliography file.

% \renewcommand*{\bibfont}{\scriptsize}
\bibliographystyle{ACM-Reference-Format}
\bibliography{taco}

%%
%% If your work has an appendix, this is the place to put it.
\appendix

\end{document}